\documentclass[runningheads]{llncs}
\usepackage[T1]{fontenc}
\usepackage{graphicx,verbatim}
\usepackage{pifont} 
\usepackage{booktabs} 
\usepackage{multirow} 
\usepackage{amsmath}
\usepackage{amssymb}  
\usepackage{hyperref}
\begin{document}
\title{Continual Retinal Vision-Language Pre-training upon Incremental Imaging Modalities}
%

\author{Yuang Yao\inst{1,2} \and
Ruiqi Wu\inst{1,2} \and
Yi Zhou\inst{1,2}\thanks{Corresponding author: Yi Zhou} \and
Tao Zhou\inst{3}}

\authorrunning{Y. Yao and Y. Zhou et al.}
%
\institute{School of Computer Science and Engineering, Southeast University, China \and Key Laboratory of New Generation Artificial Intelligence Technology and Its Interdisciplinary Applications, Ministry of Education, China \and Nanjing University of Science and Technology, China \\
\email{\{yaoyuang@seu.edu.cn, yizhou.szcn@gmail.com\}}
}

\maketitle              

\begin{abstract}
Traditional fundus image analysis models focus on single-modal tasks, ignoring fundus modality complementarity, which limits their versatility. Recently, retinal foundation models have emerged, but most still remain modality-specific. Integrating multiple fundus imaging modalities into a single foundation model is valuable. However, in dynamic environments, data from different modalities often arrive incrementally, necessitating continual pre-training. To address this, we propose RetCoP, the first continual vision-language pre-training framework in the fundus domain, which incrementally integrates image and text features from different imaging modalities into a single unified foundation model. To mitigate catastrophic forgetting in continual pre-training, we introduce a rehearsal strategy utilizing representative image-text pairs and an off-diagonal information distillation approach. The former allows the model to revisit knowledge from previous stages, while the latter explicitly preserves the alignment between image and text representations. Experiments show that RetCoP outperforms all the compared methods, achieving the best generalization and lowest forgetting rate. \textbf{The code can be found at \href{https://github.com/Yuang-Yao/RetCoP}{https://github.com/Yuang-Yao/RetCoP}}.


\end{abstract}
\section{Introduction}

Fundus imaging reveals eye and systemic diseases, enabling early diagnosis and intervention~\cite{badar2020application}. Traditional models focus on modality-specific tasks~\cite{pan2021retinal,he2019dme}, requiring separate models and limiting cross-task knowledge sharing. Recently, foundation models in fundus imaging have gained attention~\cite{du2024ret,li2024visionunite,qiu2023visionfm}. By pre-training on large-scale datasets, foundation models can provide strong initial weights for downstream tasks, enhancing transfer learning efficiency. However, most current models remain single-modality, hindering generalizability for complex clinical tasks due to insufficient multi-modal information integration.


Fundus imaging techniques are diverse, such as Color Fundus Photography (CFP), Fluorescein Fundus Angiography (FFA), and Optical Coherence Tomography (OCT), each offering diverse perspectives on retinal structure and disease. Thus, multi-modal foundational pre-training for fundus image analysis holds significant value and potential advancement. In dynamic environments, collecting all modal data at the beginning of pre-training is often impractical, as data from different imaging modalities typically arrives incrementally. Moreover, for new modalities, starting training from scratch is inefficient, necessitating continuous model updates and incremental learning during pre-training.

In continual pre-training, catastrophic forgetting—where new-stage training impairs past knowledge—is a major challenge in dynamic environments~\cite{ni2023continual}. Unlike class-incremental learning, continual pre-training lacks explicit class concepts, requiring the model to learn general representations for each fundus modality and integrate them into a unified space~\cite{zhu2023ctp}. However, later-stage features often disrupt earlier representation spaces, undermining generalization for previous modalities and worsening forgetting. Thus, efficient continual pre-training algorithms are essential.
Existing methods like MedCoSS~\cite{ye2024continual} and CMDK~\cite{tasai2025continual} use rehearsal-based self-supervised learning but rely solely on image data, neglecting paired textual descriptions. We insist that text bridges different imaging modalities. If paired image-text data can be fully utilized, it can achieve semantic understanding aligned with images and improve generalization performance. Especially in the medical domain, text descriptions enriched with expert knowledge can further improve pre-training quality.
While continual vision-language pre-training has been explored in natural images, such as CTP~\cite{zhu2023ctp} and IncCLIP~\cite{yan2022generative}, these methods focus on class-incremental tasks in natural image domains and are not directly applicable to medical modality-incremental problems.
Furthermore, recent CLIP-based continual learning methods (e.g., ZSCL~\cite{zheng2023preventing}) primarily focus on adapting pretrained models rather than continual pretraining from scratch, making them ineffective for our challenge.

To address these challenges, we propose RetCoP, short for \textbf{Ret}inal \textbf{Co}ntinual \textbf{P}re-training. To the best of our knowledge, ReCoP is the first continual vision-language pre-training framework in the retinal domain. Our framework aligns image-text pairs across incrementally arriving fundus modalities using contrastive pre-training, integrating multi-modal knowledge into a single foundation model. To mitigate catastrophic forgetting in continual training, we introduce a rehearsal strategy based on representative joint embeddings and an off-diagonal information distillation approach. The rehearsal strategy computes image-text joint embeddings using similarity-weighted representations at each stage and selects the most representative image-text pairs via K-means sampling. The off-diagonal information distillation preserves the similarity matrix from the previous stage, maintaining image-text alignment in continual contrastive learning. RetCoP outperforms all comparison methods across downstream tasks, demonstrating superior generalization and minimal forgetting across all three fundus modalities at different training stages. Our main contributions are as follows:
\begin{itemize}
  \item We propose RetCoP, the first continual vision-language pre-training framework in the fundus domain, integrating image-text pairs from mainstream modalities into a unified foundation model.
  \item We introduce a rehearsal strategy based on representative image-text pairs and an off-diagonal information distillation strategy, effectively mitigating catastrophic forgetting in continual contrastive pre-training.
  \item Experimental results show that our method outperforms all the comparison approaches, achieving the best generalization performance and the lowest forgetting rate.
\end{itemize}

\begin{figure}[t]
\includegraphics[width=\textwidth]{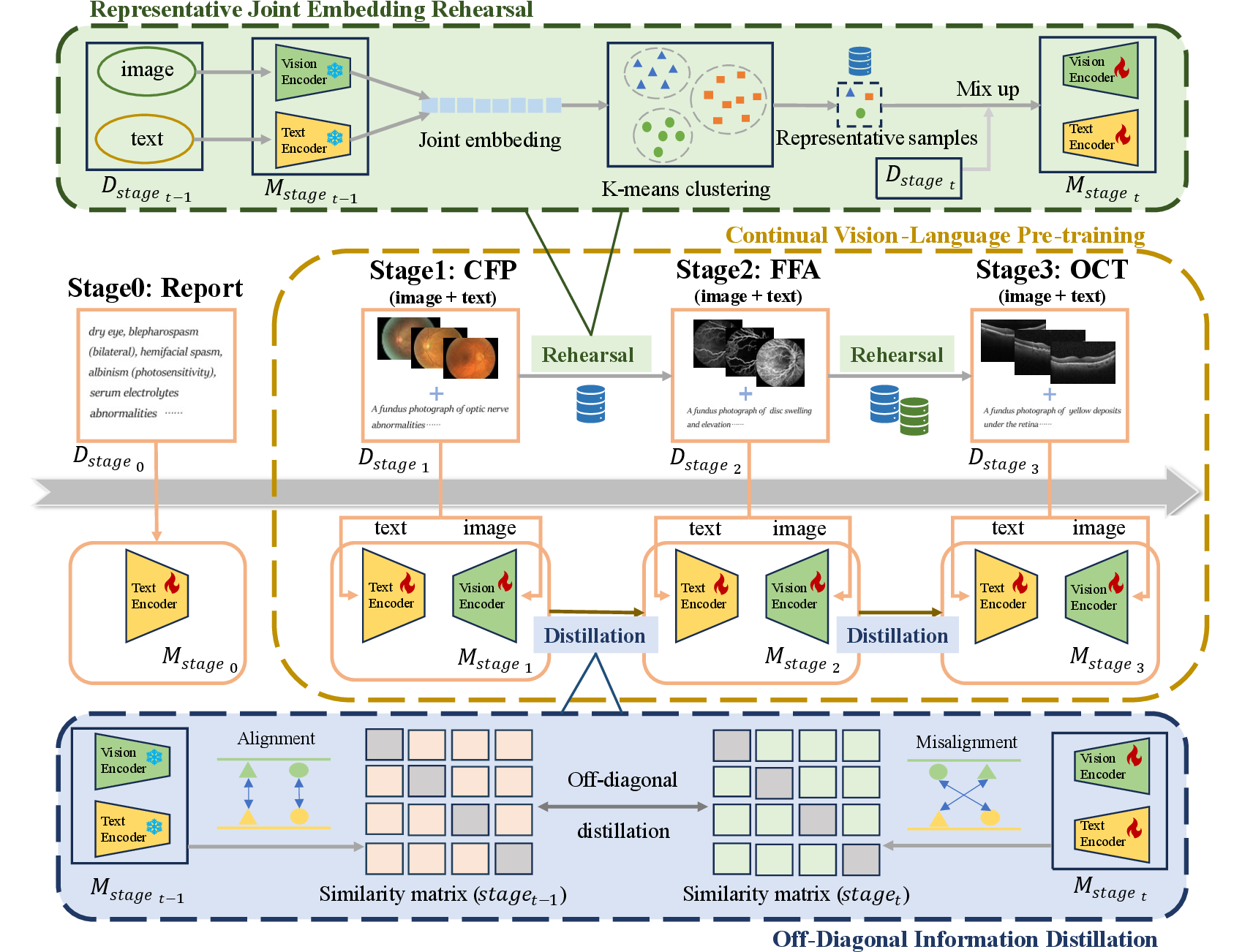}
\caption{An Overview of RetCoP. From left to right, the figure depicts the four-stage pre-training process, with stages \textsubscript{1}, \textsubscript{2}, and \textsubscript{3} performing continual vision-language pre-training on CFP, FFA, and OCT. The green section at the top shows the Representative Joint Embedding Rehearsal module, and the blue section at the bottom represents the Off-Diagonal Information Distillation module.} \label{fig1}
\end{figure}

\section{Method}
\subsection{Continual Vision-Language Pre-training Framework}
As shown in Fig.~\ref{fig1}, we propose RetCoP, a continual vision-language pre-training framework for fundus image analysis. The entire pre-training phase is divided into four stages, marked from \( \text{stage}_0 \) to \( \text{stage}_3 \). In \( \text{stage}_0 \), we pre-train the text encoder with ophthalmology and general medical report text data by masked language modeling (MLM), providing a shared semantic base across varying image modalities for better text knowledge understanding and rich semantic representations for image-text alignment in subsequent contrastive pre-training. In the subsequent three-stage continual vision-language pre-training, we use image-text pairs from the CFP, FFA, and OCT modalities, which are introduced incrementally, in accordance with their decreasing  data distribution in real clinical practice. Each stage independently learns a new modality representation.


During \( \text{stage}_1 \) to \( \text{stage}_3 \), we adopt a contrastive pre-training approach, following CLIP-like model~\cite{radford2021learning} to align images and texts. The goal is to pull positive image-text pairs closer in the representation space while pushing negative pairs further apart. Suppose that there are \(N\) image-text pairs in each batch. After encoding and projection, the image feature representations and the text feature representations are denoted as: \(I = [I_1, I_2, \dots, I_N]\) and \(T = [T_1, T_2, \dots, T_N]\). The similarity matrix \(S\) is an \(N \times N\) matrix, where each element \(S_{ij}\) represents the cosine similarity between the the feature vector \(I_i\) of \(i\)-th image and the feature vector \(T_j\) of the \(j\)-th text:
\begin{equation}
S_{ij} = \frac{I_i \cdot T_j}{\|I_i\| \|T_j\|}.
\end{equation}
The contrastive loss \( \mathcal{L}_{CLIP} \) is calculated based on the cosine similarity from both image-to-text and text-to-image. Specifically, it is formulated using the InfoNCE loss as follows:
\begin{equation}
\mathcal{L}_{CLIP} = -\frac{1}{2N} \sum_{i=1}^{N} \left[ \log \frac{\exp(S_{ii} / \tau)}{\sum_{j=1}^{N} \exp(S_{ij} / \tau)} + \log \frac{\exp(S_{ii} / \tau)}{\sum_{j=1}^{N} \exp(S_{ji} / \tau)} \right],
\end{equation}
where \( S_{ii} \) is the similarity of the positive image-text pair, \( S_{ij} \) and \( S_{ji} \) are the similarities for negative pairs, \( \tau \) is the temperature parameter.
 
In the continual pre-training scenario, an inevitable challenge is catastrophic forgetting, where knowledge acquired in previous stages is forgotten after training on a new stage. To mitigate forgetting, as shown in Fig.~\ref{fig1}, we introduce two modules into our framework: representative joint embedding rehearsal and off-diagonal information distillation, constraining the model's continual updates by revisiting representative joint embedding and preserving the alignment of representations.

\subsection{Representative Joint Embedding Rehearsal}
A primary reason for model forgetting is losing access to data from previous stages during continual pre-training. To enable the model to revisit and review past knowledge, we construct a rehearsal mechanism based on representative joint embedding, preserving a portion of image-text pairs from previous modalities on earlier stages and mixing them with the current stage's modality data (see the green section in Fig.~\ref{fig1}). 
First, to select representative samples while considering both image and text features, we compute the joint embedding representation for each image-text pair. Specifically, for image embeddings \( I_i \) and text embeddings \( T_i \), the joint embedding \( J_{i} \) is represented as:
\begin{equation}
J_{i} = S_{ii} \cdot I_{i} + (1 - S_{ii}) \cdot T_{i},
\end{equation}
where $S_{ii}$ is the similarity score between \( I_i \) and \( T_i \). If the image and text are highly aligned, the joint embedding relies more on the image information $I_{i}$; if weakly aligned, the text information $T_{i}$ is enhanced in the joint embedding to ensure the completeness of the representation.
Then, based on the computed joint embedding representations of the image-text pairs from the previous modality, we employ a k-means sampling strategy to select the most representative samples for constructing the rehearsal.
\begin{equation}
R = \bigcup_{k=1}^K \left\{ \arg \min_{i} \|J_i - c_k\| \right\}, \quad c_k = \frac{1}{|Q_k|} \sum_{i \in Q_k} J_i,
\end{equation}
where \( c_k \) is the centroid of each cluster, \( Q_k \) represents the set of samples in the \( k \)-th cluster. For each cluster, a subset of samples closest to the centroid is selected and added to the rehearsal. \( K \) is the total number of clusters, and \( R \) is the final rehearsal constructed for this stage.

In the entire continual pre-training framework, we keep a slim and fixed-size rehearsal buffer, where data from all previous modalities are evenly distributed in the rehearsal buffer. The buffer is dynamically updated during pre-training as new modalities are added, preventing it from growing too large and imposing storage and computational burdens.

\subsection{Off-Diagonal Information Distillation}
In continual vision-language pre-training, a key factor affecting forgetting rate is the drift in the alignment between textual and visual features in the representation space. The textual and visual representations learned in previous stages may experience misalignment in new stages, leading to incorrect distance relationships between paired images and texts in the representation space.

Inspired by MOD-X~\cite{ni2023continual}, to further perceive and maintain the learned alignment between images and texts while revisiting previous knowledge, we introduce an off-diagonal information distillation method(see the blue section in Fig.~\ref{fig1}). In the similarity matrix of each stage, the distribution of off-diagonal similarities reflects the current feature space. Thus, preserving off-diagonal information from past stages effectively maintains the alignment between image and text features across stages.
${S}_{t-1}$ and ${S}_{t}$ are the similarity matrices at stage$_{t-1}$ and stage$_{t}$, respectively. Since the frozen model from the previous stage may misjudge image-text alignments in the current stage, we replace rows in the ${S}_{t-1}$ (where diagonal similarity is not maximal) with corresponding values from the ${S}_{t}$ to avoid misleading updates:
\begin{equation}
S_{t-1}^{(i,:)}=S_{t}^{(i,:)};\quad \text{if}\quad \max(S_{t-1}^i)\neq i.
\end{equation}
We use KL divergence to distill the similarity distribution from the previous stage's similarity matrix, thereby explicitly constraining the updates to the representations:
\begin{equation}
\mathcal{L}_{ODID}({S}_t, {S}_{t-1}) = -\sum {S}_{t-1} \ln\left(\frac{{S}_t}{{S}_{t-1}}\right).
\end{equation}
We incorporate the distillation loss into the contrastive loss \( \mathcal{L}_{{CLIP}} \), resulting in the total loss function for RetCoP as follows, where \(\lambda\) is the weight for \(\mathcal{L}_{ODID}\):
\begin{equation}
\mathcal{L}_{{RetCoP}} = \mathcal{L}_{{CLIP}} + \lambda \cdot \mathcal{L}_{{ODID}}.
\end{equation}

\section{Experiments}
\subsection{Experimental Setup}
\noindent\textbf{Pre-training Data.}
The text report in \( \text{stage}_0 \) is sourced from 34 public medical text datasets, spanning ophthalmology and general medicine~\cite{wu2025mm}.
From \( \text{stage}_1 \) to \( \text{stage}_3 \), we conduct continual vision-language pretraining using image-text pairs.
For CFP modality, we use MM-Retinal\cite{wu2024mm} and flair dataset~\cite{silva2025foundation}, which integrates 37 public fundus datasets covering 96 categories, totaling 193,678 pairs.
For FFA modality, along with MM-Retinal\cite{wu2024mm}, we utilize a subset of the FFA-IR~\cite{li2021ffa}, which covers 46 retinal diseases, resulting in a combined total of 701,098 pairs.
For OCT modality, eleven public category-labeled OCT datasets and MM-Retinal~\cite{wu2025mm} are leveraged, providing 183,817 pairs.
Note that the categorical labels are mapped into textual prompts using prior knowledge.

\noindent\textbf{Evaluation Data and Metrics.}
Disease classification was conducted on five multi-class fundus downstream datasets across CFP, FFA, and OCT modalities under zero-shot, linear probe, and CLIP-adapter~\cite{gao2024clip} settings. 
These datasets cover many common retinal diseases like diabetic retinopathy, glaucoma, and age-related macular degeneration, among others. 
For CFP modality, the model was never trained on FIVES~\cite{jin2022fives} but partially on ODIR~\cite{odir2019} with three unseen categories withheld.
Current-stage results demonstrate model plasticity, while previous-stage results reflect forgetting degree.

\noindent\textbf{Methods for Comparison.}
RetCoP is compared with five popular continual learning methods: 1) SeqFT: The model is fine-tuned sequentially per stage. 2) LWF~\cite{li2017learning}: a regularization-based method using cross-entropy distillation. 3) EWC~\cite{kirkpatrick2017overcoming}: a regularization-based method leveraging Fisher matrix for weight constraints. 4) ER~\cite{chaudhry2019continual}: a replay-based method with reservoir sampling. 5) ICARL ~\cite{rebuffi2017icarl}: a replay-based method applying MoF sampling based on feature means.

\noindent\textbf{Implementation Details.}
We employed ResNet-50~\cite{he2016deep} initialized with ImageNet-1K as the vision encoder and BioClinicalBERT~\cite{alsentzer2019publicly} as the text encoder. 
Images are resized to 512×512, and English text tokens are set to a length of 256. 
OCT images are resized with isotropic scaling and zero-padding.
The evaluation uses five-fold cross-validation with averaged results. 
Training was conducted using the AdamW optimizer with a warm-up strategy on four NVIDIA A6000 GPUs, with a batch size of 24. The CLIP-like training follows~\cite{wu2024mm}.

\begin{table}[htbp]
\centering
\caption{Performance Comparison on CFP Modality (FIVES and ODIR Datasets). Training \(\text{FFA}\) and \(\text{OCT}\) in stages \textsubscript{2} and \textsubscript{3} induces forgetting in \(\text{CFP}\). The table shows downstream test results on CFP after all three stages, with forgetting rate relative to stage \textsubscript{1} in parentheses for stages \textsubscript{2} and \textsubscript{3}.}
\fontsize{8pt}{9.6pt}\selectfont
\begin{tabular}{c|c|cc|cc|cc}
\cline{1-8}
\multicolumn{8}{l}{\textbf{FIVES Dataset}} \\ \cline{1-8}
\multirow{2}{*}{Stage} & \multirow{2}{*}{Method} & \multicolumn{2}{c}{Zero-Shot} & \multicolumn{2}{c}{Linear Probe} & \multicolumn{2}{c}{ClipAdapter} \\ \cline{3-8}
                       &                         & ACC(\%) & AUC(\%) & ACC(\%) & AUC(\%) & ACC(\%) & AUC(\%) \\ \cline{1-8}
1                      & -                       & 58.9    & 88.6    & 82.6    & 96.1    & 81.6    & 95.0    \\ \cline{1-8}
\multirow{6}{*}{2}     & RetCoP                  & \textbf{52.0(↓6.9)} & \textbf{82.2(↓6.4)} & \textbf{79.7(↓2.9)} & \textbf{94.5(↓1.6)} & \textbf{73.7(↓7.9)} & \textbf{91.7(↓3.3)} \\
                       & SeqFT                   & 27.5(↓31.4) & 60.7(↓27.9) & 63.1(↓19.5) & 85.2(↓10.9) & 57.1(↓24.5) & 82.7(↓12.3) \\
                       & LWF                     & 44.2(↓14.7) & 72.5(↓16.1) & 72.4(↓10.2) & 90.9(↓5.2) & 64.6(↓17.0) & 87.1(↓7.9) \\
                       & EWC                     & 33.1(↓25.8) & 65.6(↓23.0) & 70.6(↓12.0) & 89.5(↓6.6) & 62.5(↓19.1) & 84.2(↓10.8) \\
                       & ICARL                   & 39.4(↓19.5) & 65.0(↓23.6) & 73.2(↓9.4) & 90.9(↓5.2) & 64.8(↓16.8) & 86.5(↓8.5) \\
                       & ER                      & 30.0(↓28.9) & 74.5(↓14.1) & 74.7(↓7.9) & 92.7(↓3.4) & 61.2(↓20.4) & 85.9(↓9.1) \\ \cline{1-8}
\multirow{6}{*}{3}     & RetCoP                  & \textbf{45.9(↓13.0)} & \textbf{76.7(↓11.9)} & \textbf{77.6(↓5.0)} & \textbf{93.7(↓2.4)} & \textbf{72.3(↓9.3)} & \textbf{91.0(↓4.0)} \\
                       & SeqFT                   & 21.9(↓37.0) & 39.5(↓49.1) & 57.4(↓25.2) & 83.6(↓12.5) & 55.6(↓26.0) & 80.7(↓14.3) \\
                       & LWF                     & 29.5(↓29.4) & 59.3(↓29.3) & 68.3(↓14.3) & 88.1(↓8.0) & 42.6(↓39.0) & 70.7(↓24.3) \\
                       & EWC                     & 21.5(↓37.4) & 49.9(↓38.7) & 64.0(↓18.6) & 86.3(↓9.8) & 48.7(↓32.9) & 77.8(↓17.2) \\
                       & ICARL                   & 29.2(↓29.7) & 57.8(↓30.8) & 66.4(↓16.2) & 87.0(↓9.1) & 39.7(↓41.9) & 68.5(↓26.5) \\
                       & ER                      & 26.1(↓32.8) & 56.3(↓32.3) & 73.5(↓9.1) & 91.4(↓4.7) & 62.1(↓19.5) & 85.4(↓9.6) \\ \cline{1-8}

\multicolumn{8}{l}{\textbf{ODIR Dataset}} \\ \cline{1-8}
\multirow{2}{*}{Stage} & \multirow{2}{*}{Method} & \multicolumn{2}{c}{Zero-Shot} & \multicolumn{2}{c}{Linear Probe} & \multicolumn{2}{c}{ClipAdapter} \\ \cline{3-8}
                       &                         & ACC(\%) & AUC(\%) & ACC(\%) & AUC(\%) & ACC(\%) & AUC(\%) \\ \cline{1-8}
1                      & -                       & 80.0    & 93.8    & 89.0    & 97.9    & 89.0    & 96.6    \\ \cline{1-8}
\multirow{6}{*}{2}     & RetCoP                  & \textbf{68.8(↓11.2)} & \textbf{90.0(↓3.8)} & \textbf{89.8(↑0.8)} & \textbf{97.9(0.0)} & \textbf{89.1(↑0.1)} & \textbf{97.7(↑1.1)} \\
                       & SeqFT                   & 29.8(↓50.2) & 48.2(↓45.6) & \textbf{89.8(↑0.8)} & 96.4(↓1.5) & 87.7(↓1.3) & 96.0(↓0.6) \\
                       & LWF                     & 41.5(↓38.5) & 57.8(↓36.0) & 89.0(0.0) & 96.9(↓1.0) & 87.5(↓1.5) & 96.3(↓0.3) \\
                       & EWC                     & 36.5(↓43.5) & 70.7(↓23.1) & 88.3(↓0.7) & 96.0(↓1.9) & 86.0(↓3.0) & 95.3(↓1.3) \\
                       & ICARL                   & 59.8(↓20.2) & 85.8(↓8.0) & 89.0(0.0) & 97.1(↓0.8) & 88.2(↓0.8) & 96.5(↓0.1) \\
                       & ER                      & 53.0(↓27.0) & 73.4(↓20.4) & 88.3(↓0.7) & 96.6(↓1.3) & 87.5(↓1.5) & 96.8(↑0.2) \\ \cline{1-8}
\multirow{6}{*}{3}     & RetCoP                  & \textbf{56.2(↓23.8)} & \textbf{81.5(↓12.3)} & \textbf{91.3(↑2.3)} & \textbf{98.0(↑0.1)} & \textbf{90.7(↑1.7)} & \textbf{97.3(↑0.7)} \\
                       & SeqFT                   & 30.0(↓50.0) & 47.1(↓46.7) & 73.5(↓15.5) & 88.4(↓9.5) & 69.3(↓19.7) & 85.2(↓11.4) \\
                       & LWF                     & 35.7(↓44.3) & 51.9(↓41.9) & 82.0(↓7.0) & 92.0(↓5.9) & 66.5(↓22.5) & 81.9(↓14.7) \\
                       & EWC                     & 32.7(↓47.3) & 51.3(↓42.5) & 77.3(↓11.7) & 90.3(↓7.6) & 75.8(↓13.2) & 90.5(↓6.1) \\
                       & ICARL                   & 34.3(↓45.7) & 56.3(↓37.5) & 79.8(↓9.2) & 91.6(↓6.3) & 66.8(↓22.2) & 82.2(↓14.4) \\
                       & ER                      & 38.8(↓41.2) & 67.4(↓26.4) & 88.7(↓0.3) & 95.8(↓2.1) & 87.7(↓1.3) & 95.9(↓0.7) \\ \cline{1-8}
\end{tabular}
\label{tab:merged_cfp_results}
\end{table}

\begin{table}[htbp]
\centering
\caption{Performance Comparison on FFA Modality (MPOS Dataset). Stage \textsubscript{3} trains OCT, inducing FFA forgetting. The table shows downstream test results on FFA for stages \textsubscript{2} and \textsubscript{3}, with values in parentheses indicating forgetting rate relative to stage \textsubscript{2}.} 
\fontsize{8pt}{9.6pt}\selectfont 
\begin{tabular}{cc|cc|cc|cc}
\hline
\multicolumn{1}{c|}{\multirow{2}{*}{Stage}} & \multicolumn{1}{c|}{\multirow{2}{*}{Method}} & \multicolumn{2}{c|}{Zero-Shot} & \multicolumn{2}{c|}{Linear Probe} & \multicolumn{2}{c}{ClipAdapter} \\ \cline{3-8} 
\multicolumn{1}{c|}{}                       & \multicolumn{1}{c|}{}                        & ACC(\%) & AUC(\%) & ACC(\%) & AUC(\%) & ACC(\%) & AUC(\%) \\ \hline
\multicolumn{1}{c|}{\multirow{6}{*}{2}}     & \multicolumn{1}{c|}{RetCoP}                  & \textbf{49.8} & \textbf{88.0} & \textbf{86.2} & \textbf{97.7} & 80.7 & \textbf{96.8} \\
\multicolumn{1}{c|}{}                       & \multicolumn{1}{c|}{SeqFT}                   & 29.6 & 81.9 & 83.5 & 96.8 & 73.6 & 94.7 \\
\multicolumn{1}{c|}{}                       & \multicolumn{1}{c|}{LWF}                     & 44.7 & 80.4 & 85.5 & 97.2 & \textbf{82.5} & 96.5 \\
\multicolumn{1}{c|}{}                       & \multicolumn{1}{c|}{EWC}                     & 36.0 & 74.3 & 84.2 & 96.9 & 74.8 & 95.3 \\
\multicolumn{1}{c|}{}                       & \multicolumn{1}{c|}{ICARL}                   & 37.8 & 76.2 & 77.9 & 95.1 & 76.7 & 94.1 \\
\multicolumn{1}{c|}{}                       & \multicolumn{1}{c|}{ER}                      & 49.4 & 86.3 & 80.7 & 95.9 & 72.1 & 94.2 \\ \hline
\multicolumn{1}{c|}{\multirow{6}{*}{3}}     & \multicolumn{1}{c|}{RetCoP}                  & \textbf{25.4(↓24.4)} & \textbf{70.5(↓17.5)} & \textbf{83.0(↓3.2)} & \textbf{96.6(↓1.1)} & 76.6(↓4.1) & 95.0(↓1.8) \\
\multicolumn{1}{c|}{}                       & \multicolumn{1}{c|}{SeqFT}                   & 18.0(↓11.6) & 48.8(↓33.1) & 76.9(↓6.6) & 94.1(↓2.7) & 54.0(↓19.6) & 87.9(↓6.8) \\
\multicolumn{1}{c|}{}                       & \multicolumn{1}{c|}{LWF}                     & 21.7(↓23.0) & 69.8(↓11.1) & 80.7(↓4.8) & 95.5(↓1.7) & 62.3(↓20.2) & 92.5(↓2.2) \\
\multicolumn{1}{c|}{}                       & \multicolumn{1}{c|}{EWC}                     & 17.9(↓18.1) & 49.5(↓24.8) & 70.8(↓13.4) & 93.5(↓3.4) & 50.7(↓24.1) & 84.7(↓10.6) \\
\multicolumn{1}{c|}{}                       & \multicolumn{1}{c|}{ICARL}                   & 20.8(↓17.0) & 60.6(↓15.6) & 77.7(↓0.2) & 95.5(↓0.4) & 67.2(↓5.5) & 92.9(↓1.2) \\
\multicolumn{1}{c|}{}                       & \multicolumn{1}{c|}{ER}                      & 23.2(↓26.2) & 63.9(↓22.4) & 81.1(↓0.4) & 96.1(↓0.1) & \textbf{77.9(↑5.8)} & \textbf{95.3(↑1.1)} \\ \hline
\end{tabular}
\label{tab:mpos_ffa_results} 
\end{table}

\subsection{Experimental Results and Analysis}
\noindent\textbf{Method Comparison.}
From Table~\ref{tab:merged_cfp_results}, it is clear that RetCoP achieves the lowest forgetting rates and maintains superior performance across all stages and metrics in CFP modality, especially in zero-shot. For instance, on FIVES dataset~\cite{jin2022fives}, RetCoP shows only \textbf{6.9\%} zero-shot ACC drop in \( \text{stage}_2 \) (vs. \textbf{31.4\%} for SeqFT) and \textbf{13.0\%} in \( \text{stage}_3 \) (vs. \textbf{37.0\%} for SeqFT), demonstrating strong resistance to catastrophic forgetting. Notably, after training on FFA \( \text{stage}_2 \) and OCT \( \text{stage}_3 \), RetCoP retains \textbf{45.9\%} zero-shot ACC for CFP in \( \text{stage}_3 \), while others (e.g., LWF: \textbf{29.5\%}, EWC: \textbf{21.5\%}) suffer severe degradation. On ODIR dataset~\cite{odir2019}, RetCoP not only mitigates forgetting (e.g., \( \text{stage}_3 \) zero-shot AUC: $\downarrow$\textbf{12.3\%} vs. SeqFT's $\downarrow$\textbf{46.7\%}) but also even improves some metrics (e.g., ClipAdapter AUC: $\uparrow$\textbf{1.1\%} in \( \text{stage}_2 \)).

As shown in Table~\ref{tab:mpos_ffa_results}, on the MPOS dataset~\cite{wang2024non} (FFA modality), RetCoP achieves the least forgetting in $\text{stage}_3$, with only \textbf{3.2\%} ACC decline in linear probe and \textbf{1.1\%} AUC drop -- significantly lower than others like SeqFT (\textbf{$\downarrow$6.6\%} ACC/$\downarrow$2.7\% AUC) and EWC (\textbf{$\downarrow$13.4\%} ACC/$\downarrow$3.4\% AUC). While ER shows accidental improvement in ClipAdapter ($\uparrow$5.8\% ACC), its severe forgetting in zero-shot (\textbf{$\downarrow$26.2\%} ACC) reveals instability. RetCoP achieves these results by effectively balancing cross-modal knowledge through rehearsal and distillation, ensuring stable performance on prior tasks while adapting to new modalities.
\begin{table}[t]
\centering
\caption{Performance Comparison on OCT Modality (ACC \& AUC, \%). The data in the table are averaged under zero-shot, linear
probe and clip-adapter settings.}
\label{tab:oct_compact}
\fontsize{8pt}{8pt}\selectfont
\begin{tabular}{@{}lcccccc@{}}
\hline
\multirow{2}{*}{Dataset} & \multicolumn{6}{c}{Method} \\
\cmidrule(lr){2-7}
 & RetCoP & SeqFT & LWF & EWC & ICARL & ER \\ 
\midrule
OCTID (ACC/AUC) & \textbf{71.6}/94.6 & 70.5/94.2 & 69.9/93.4 & 69.7/\textbf{95.5} & 52.1/90.4 & 67.5/92.5 \\
OCTDL (ACC/AUC) & \textbf{83.1}/\textbf{88.1} & 82.4/87.4 & 81.7/83.8 & 82.6/87.7 & 71.3/86.8 & 80.0/84.0 \\
\hline
\end{tabular}
\end{table}
The results in FFA (Table~\ref{tab:mpos_ffa_results}) and OCT (Table~\ref{tab:oct_compact}, on OCTID~\cite{gholami2020octid} and OCTDL~\cite{kulyabin2024octdl} datasets) experiments demonstrate that RetCoP not only effectively mitigates forgetting but also achieves the best performance in learning FFA and OCT modalities during stage\textsubscript{2} and stage\textsubscript{3}, showcasing strong plasticity.
\begin{table}[t]
\centering
\caption{Ablation Study. The data in the table are averaged under zero-shot, linear probe and clip-adapter settings.} 
\fontsize{8pt}{9.6pt}\selectfont 
\begin{tabular}{ccc|cc|cc|cc}
\hline
\multicolumn{1}{c}{\multirow{2}{*}{Rehearsal}} & \multirow{2}{*}{ODID} & \multirow{2}{*}{Re-init} & \multicolumn{2}{c|}{CFP\_stage2} & \multicolumn{2}{c|}{CFP\_stage3} & \multicolumn{2}{c}{FFA\_stage3} \\ \cline{4-9}
                        &                       &                          & ACC(\%) & AUC(\%) & ACC(\%) & AUC(\%) & ACC(\%) & AUC(\%) \\ \hline
$\checkmark$               & $\checkmark$          & $\checkmark$             & 71.3 & 89.3 & 64.1 & 86.4 & 58.7 & 85.7 \\
                          & $\checkmark$          &                         & 62.8 & 81.2 & 51.6 & 73.7 & 56.5 & 83.5 \\
$\checkmark$               &                      &                        & 69.3 & 84.6 & 65.6 & 88.1 & 57.4 & 86.3 \\
(ours)  $\checkmark$             & $\checkmark$          &                         & \textbf{75.5} & \textbf{92.3} & \textbf{72.3} & \textbf{89.7} & \textbf{61.7} & \textbf{87.4} \\ \hline
\end{tabular}
\label{tab:ablation} 
\end{table}

\noindent\textbf{Ablation Studies.}
To validate the effectiveness of each module in mitigating catastrophic forgetting, we designed ablation experiments using the control variable method (Table \ref{tab:ablation}). The results show that removing the rehearsal or distillation component increases the model's average forgetting rate in CFP and FFA modalities, highlighting their critical role in facilitating knowledge retention and mitigating catastrophic forgetting. Additionally, we found that reinitializing the text encoder with \( \text{stage}_0 \) weights during incremental pre-training led to higher forgetting rates than using the previous stage's weights.

\section{Conclusion}
In this paper, we propose RetCoP, the first continual pretraining framework in fundus domain, learning incremental imaging modalities. Experiments demonstrate that, by designing the rehearsal strategy and off-diagonal information distillation, RetCoP effectively achieves optimal performance and mitigates catastrophic forgetting during continual pre-training. We hope this work will inspire further research for incremental pretraining in dynamic clinical environments.

\begin{credits}
\subsubsection{\ackname} This work was partially supported by the National Natural Science Foundation of China (Nos. 62476054, and 62172228), and the Fundamental Research Funds for the Central Universities of China. 
This preprint version is prior to peer review. The Version of Record of this contribution can be referred to proceedings of MICCAI 2025.

\end{credits}

%
%
%
\bibliographystyle{splncs04}
\bibliography{reference.bib}
%

\end{document}